\documentclass[runningheads]{llncs}
\usepackage{graphicx}
\usepackage{comment}
\usepackage{amsmath,amssymb} 
\usepackage{color} 
\usepackage{xspace}

\usepackage{epstopdf}

\usepackage{times}
\usepackage{epsfig}

\usepackage{bm}
\usepackage{bbm}
\newcommand{\be}{\begin{equation}}
\newcommand{\ee}{\end{equation}}
\usepackage{amsmath}
\DeclareMathOperator*{\argmin}{\arg\!\min}
\DeclareMathOperator*{\argmax}{\arg\!\max}

\makeatletter
\DeclareRobustCommand\onedot{\futurelet\@let@token\@onedot}
\def\@onedot{\ifx\@let@token.\else.\null\fi\xspace}

\def\eg{\emph{e.g}\onedot} 
\def\ie{\emph{i.e}\onedot}

\def\etal{\emph{et al}\onedot}

\begin{document}
\pagestyle{headings}
\mainmatter
\def\ECCVSubNumber{}  

\title{From Real to Synthetic and Back: Synthesizing Training Data for Multi-Person Scene Understanding} 

\titlerunning{ECCV-20 submission ID \ECCVSubNumber} 
\authorrunning{ECCV-20 submission ID \ECCVSubNumber} 
\author{Anonymous ECCV submission}
\institute{Paper ID \ECCVSubNumber}

\titlerunning{I. Kviatkovsky et al.: From Real to Synthetic and Back: Synthesizing Humans}
%
\author{Igor Kviatkovsky  \and
Nadav Bhonker \and
Gerard Medioni}
\authorrunning{I. Kviatkovsky, N. Bhonker, G. Medioni}
%
\institute{Amazon Go \\ \email{\{kviat,~nadavb,~medioni\}@amazon.com}}
\maketitle

\begin{abstract}
    We present a method for synthesizing naturally looking images of multiple people interacting in a specific scenario. 
    These images benefit from the advantages of synthetic data: being fully controllable and fully annotated with any type of standard or custom-defined ground truth.
    To reduce the synthetic-to-real domain gap, we introduce a pipeline consisting of the following steps: 1) we render scenes in a context modeled after the real world, 2) we train a human parsing model on the synthetic images, 3) we use the model to estimate segmentation maps for real images, 4) we train a conditional generative adversarial network (cGAN) to learn the inverse mapping -- from a segmentation map to a real image, and 5) given new synthetic segmentation maps, we use the cGAN to generate realistic images.
    An illustration of our pipeline is presented in Figure~\ref{fig:full_pipeline}.
    We use the generated data to train a multi-task model on the challenging tasks of UV mapping and dense depth estimation.
    We demonstrate the value of the data generation and the trained model, both quantitatively and qualitatively on the CMU Panoptic Dataset.
\end{abstract}

\section{Introduction}
    \begin{figure}[tb]
    \begin{center}
    \begin{tabular}{cc}
    \includegraphics[width=5.35cm]{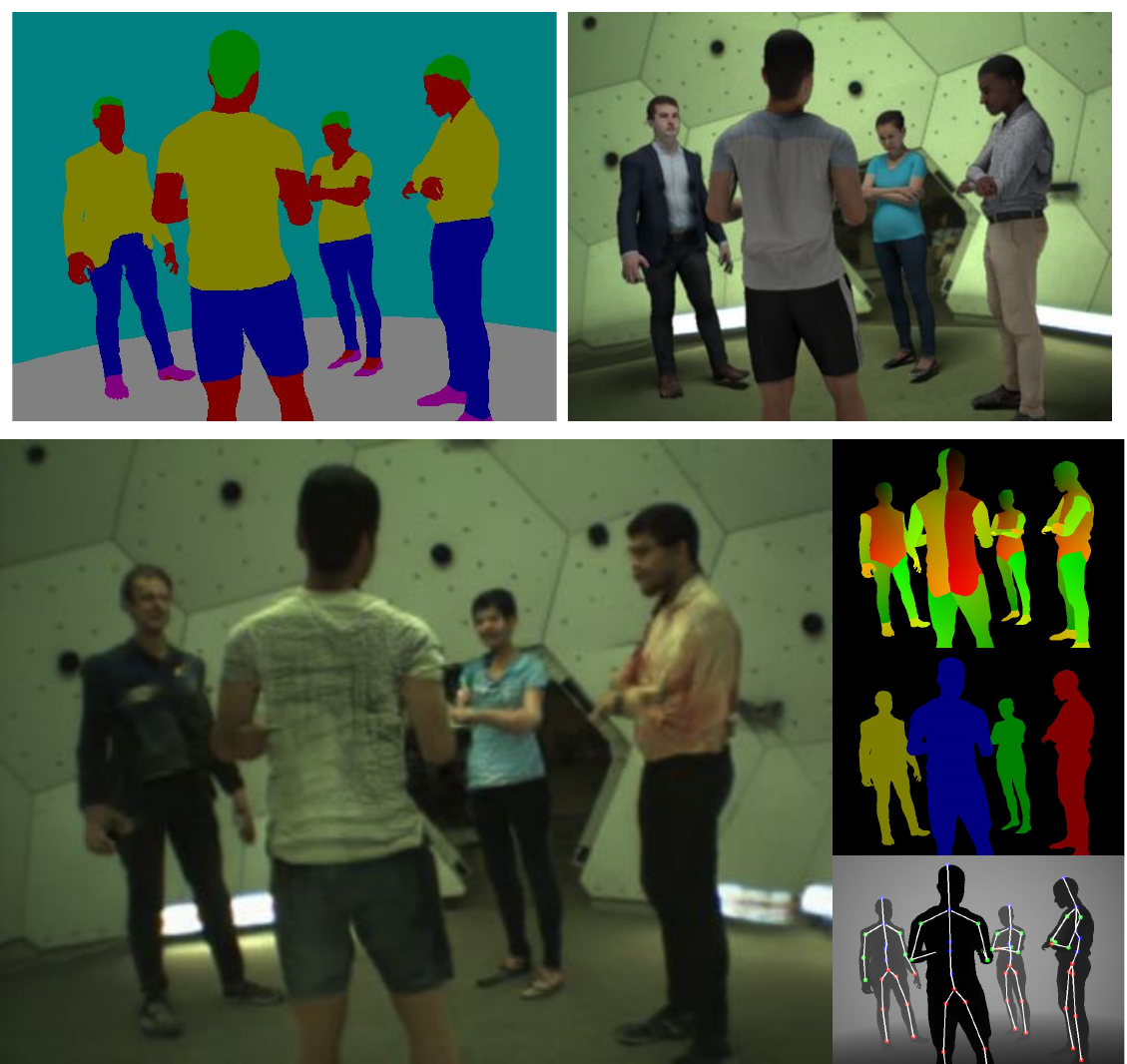} &
 \includegraphics[width=6.65cm]{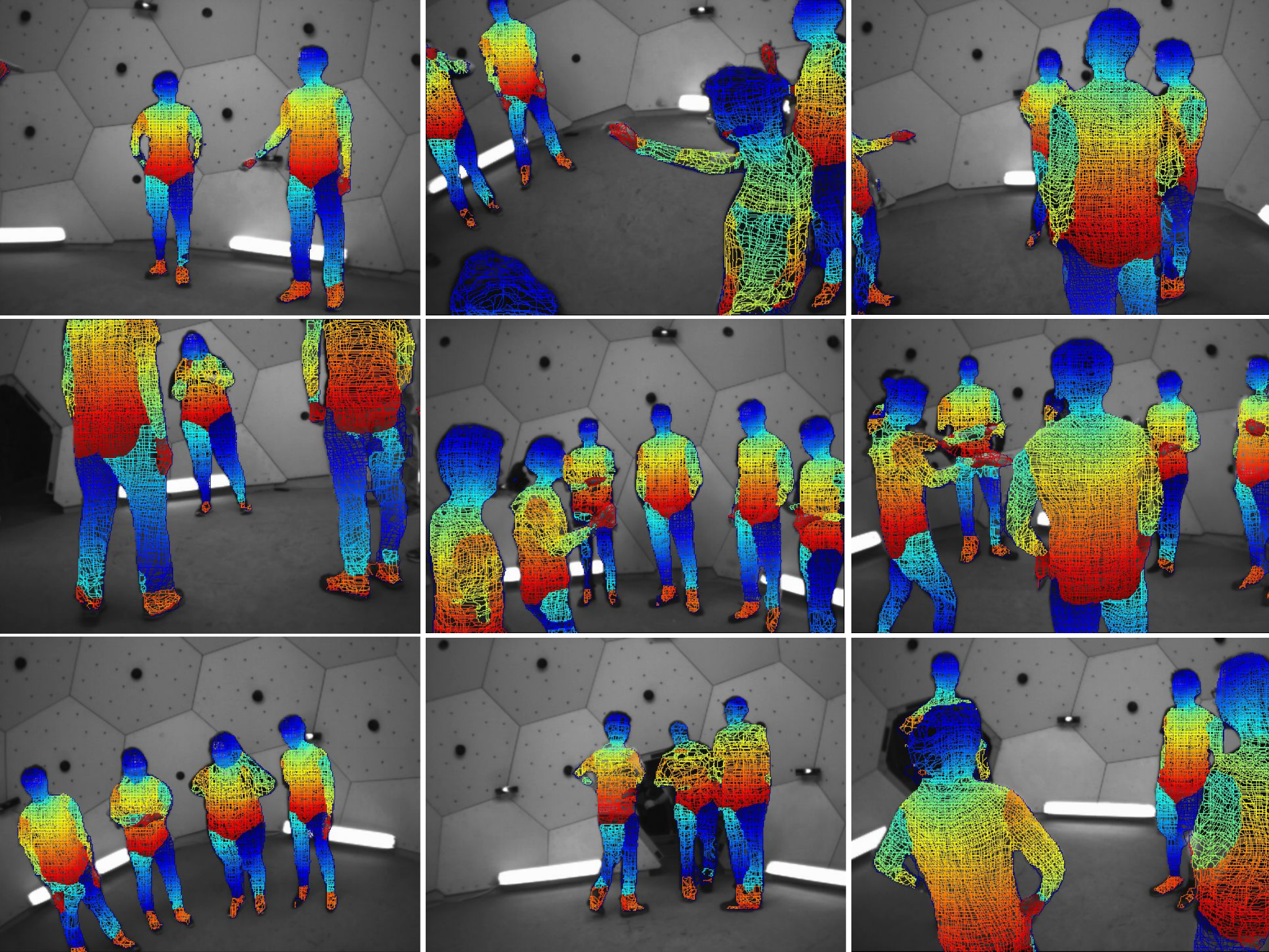} \\
     (a) & (b)
    \end{tabular} 
    \caption{(a) Bottom: image generated by our system pre-conditioned on the appearance segmentation map and the rendered frame (top row) defining the layout of the scene and the person appearances. To the right of the generated frame three examples of automatically generated ground truth annotations: UV mapping, instance segmentation, depth and skeleton. (b) Real CMU Panoptic frames with the inferred UV maps superimposed.}
    \label{fig:output}
    \end{center}
    \end{figure}

    Humans are by far the most widely studied topic in the computer vision community.
    This increased interest occurred naturally due to the many human-related applications that utilize images.
    These applications range from ``low level'' tasks such as face detection, pedestrian detection to higher-level tasks such as pedestrian tracking, body pose estimation and human action recognition.
    Recently, the field has seen a great leap in performance and abilities due to the incorporation of deep neural networks (DNNs).
    Unfortunately, for most applications, DNNs require large amounts of annotated data. 
    Acquiring adequate amounts of data is a tedious and expensive endeavor. 
    To mitigate costs several solutions were proposed.
    One such solution is auto-annotation -- automatically annotate data using either a single powerful off-the-shelf model (a.k.a. oracle) or an ensemble of models~\cite{cheng2018survey,simon2019autoannotation}.
    Although auto-annotation significantly reduces the amount of required manual annotation, it does not reduce the need for the data collection itself each time a system-level change is introduced, \eg, novel scene, camera view, object appearances, etc.
    Synthetic data is a powerful and cost-effective solution decoupling the trained model from the physical world.
    The idea is to synthesize photorealistic images with automatically generated annotations~\cite{gaidon2016virtual,varol17_surreal,sankaranarayanan2018learning,chen2016synthesizing}.
    Additional benefits of synthetic data are its full interpretability and the absence of annotation errors and inconsistencies that inevitably arise in auto-annotated and manually annotated data.
    As the amount of synthetic data increases in the training set, the real-to-synthetic domain gap becomes a more central barrier for efficient generalization.
    The gap can be decomposed into two independent factors -- appearance and content.
    The \textit{appearance domain gap} arises due to the difference between the natural and the synthetic image formation processes.
    A real image is a product of an image acquisition process, inherently captures a variety of different natural phenomena, such as motion noise and material properties.
    On the other hand, a synthetic image is the output of light transport simulation and rendering.
    Although state-of-the-art computer graphics technologies have reached the point where simulation is almost indistinguishable from real imagery, they require significant investments in 3D modeling and rendering infrastructures.
    An alternative approach is to lower the quality bar and leverage the available images from the target domain to close this gap~\cite{wang2018high,zhu2017unpaired}.
    The \textit{content domain gap}~\cite{dvornik2018modeling,kar2019meta,tripathi2019learning} is a result of another fundamental difference between the domains, arising from the differences in object types and their layouts.
    For instance, both datasets for autonomous driving and fashion contain people. 
    However, their poses, viewpoints, illumination, clothing variety, etc. are widely different.
    To successfully bridge between two domains, one has to address both factors responsible for the gap.

    The effectiveness of synthetic data has been demonstrated in the field of self-driving cars following the introduction of several synthetic datasets~\cite{Dosovitskiy17,gaidon2016virtual,prakash2018structured,richter2017playing}.
    Recently, several human-centric synthetic datasets were introduced for person re-identification~\cite{bak2018domain,barbosa2018looking,ma2017pose,qian2018pose}, pose estimation~\cite{ghezelghieh2016learning,hoffmann2019learning,lassner2017generative} and body-part segmentation~\cite{lassner2017generative,varol17_surreal}
    Most of these datasets were generated using either classic 3D graphics techniques~\cite{barbosa2018looking,ghezelghieh2016learning,hoffmann2019learning,varol17_surreal} or generative adversarial networks (GANs)~\cite{goodfellow2014generative} trained to generate 2D photorealistic frames directly in image space~\cite{ma2017pose,qian2018pose}.
    Some works use a combination of these two approaches~\cite{bak2018domain,lassner2017generative}.
    The former approach incorporates 3D geometric priors related to human body pose and appearance, while the latter is used to reduce the appearance domain gap.
    Typically, a single person was considered, \eg, re-id, fashion, and blank or random backgrounds were used for achieving resiliency to different scene types.
    Due to the targeted applications' nature, modeling inter-personal and environmental contexts was not considered by these works.
    
    We introduce a methodology for synthesizing naturally looking images of multiple people interacting in a specific scenario.
    These images benefit from the advantages of synthetic data: being fully controllable and fully annotated with any type of standard or custom-defined ground truth, \eg, semantic and instance labels, UV maps, depth measurements, body joint locations and fiducial keypoint locations.
    We leverage 3D computer graphics and conditional GANs to reduce the content and appearance domain gaps, respectively.
    First, we construct a 3D virtual space mimicking both the environment and the humans from the target domain, and use it to generate synthetic frames of multiple interacting individuals, annotated with appearance segmentation (a.k.a. human parsing) labels.
    \begin{figure}[tb]
    \begin{center}
    \begin{tabular}{cc}
    \includegraphics[width=6cm]{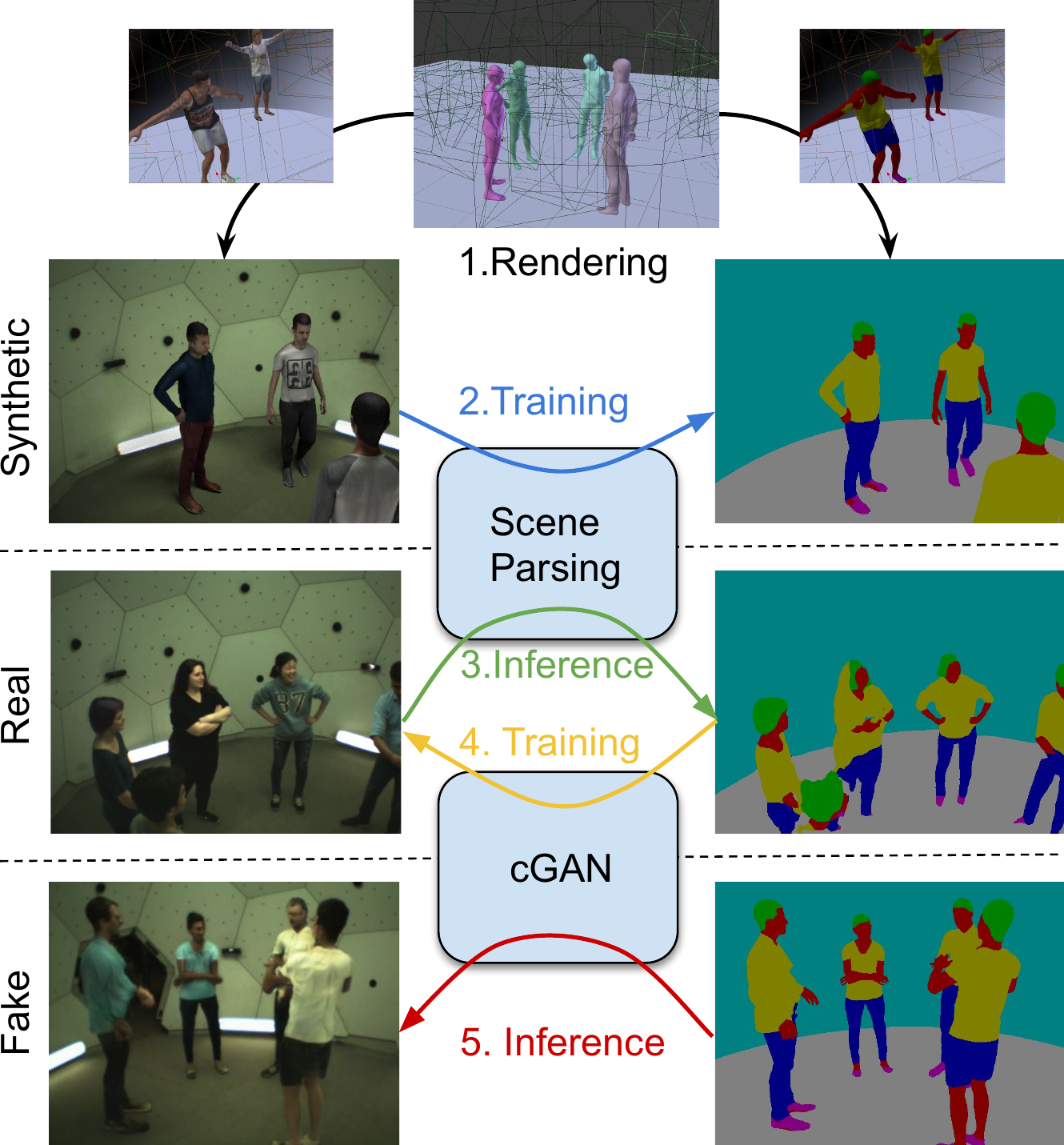} &
    \includegraphics[width=6cm]{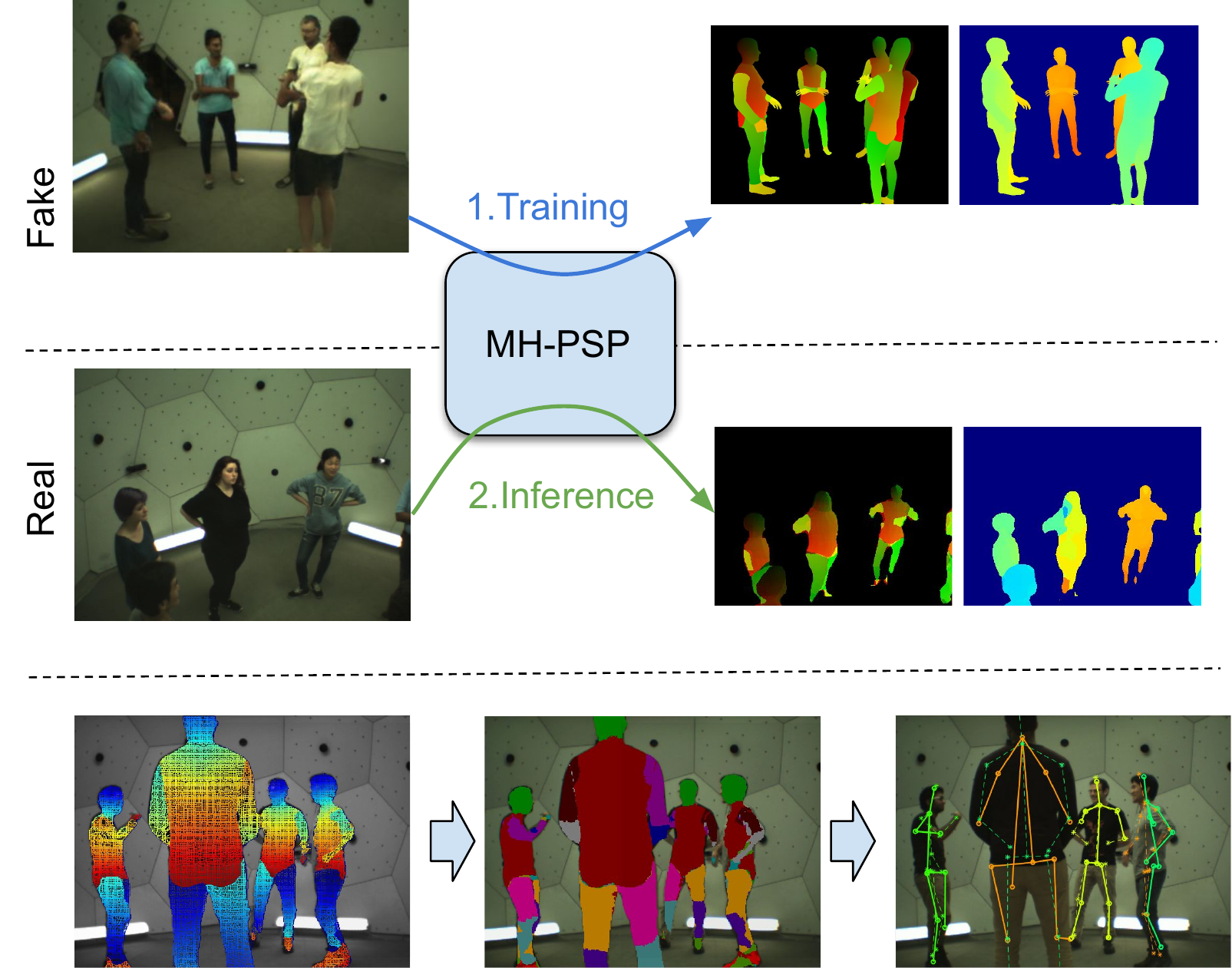} \\
    (a) & (b)
    \end{tabular}
    \caption{(a) The five main stages of our data generation pipeline: 1) Scene 3D modeling and frame rendering. 2) Training the scene parsing model on the rendered frames. 3) Real domain auto-annotation with the scene parsing model from stage 2. 4) Training SPADE on the auto-annotated data. 5) Generate photorealistic images from synthetic scene parsing annotations.
    (b) The narrowed domain gap allows us now to train a new model for UV mapping and depth estimation and apply it on real images. The inferred UV mapping can be used for part segmentation and 2D pose estimation.}
    \label{fig:full_pipeline}
    \end{center}
    \end{figure}
    Second, we train a scene parsing model~\cite{zhao2017pyramid} using images generated in the target context with additional real images in the wild~\cite{zhao2018understanding}.
    Third, we use the model to auto-annotate frames from the target domain.
    Finally, we train a cGAN (based on~\cite{park2019semantic} and~\cite{wang2018high}) to generate a photo realistic image from an appearance segmentation label map.
    This allows us to generate a large variety of photorealistic frames by rendering semantic segmentation maps corresponding to a specific 3D scene and transforming them via the cGAN.
    An illustration of stages of our proposed method applied to the CMU Panoptic domain~\cite{Joo_2017_TPAMI} is presented in Figure~\ref{fig:full_pipeline}.
    We demonstrate the usefulness of this data by learning to predict both UV mapping and depth maps.
    Acquiring these ground truth annotations otherwise is very expensive, requiring either dedicated hardware or an elaborate annotation process~\cite{guler2018densepose}.
    We emphasize that we cross the domain gap three times.
    First, we cross from synthetic to real by training a human parsing model on synthetic images and apply it to real images from the target domain.
    Second, we train a generative model on real images for the opposite task -- to create a realistic image from a synthesized semantic segmentation map.
    Third, we train a semantic segmentation model on these fake realistic images and infer on real images.
    Below is a list summarizing our main contributions in this work:
    \begin{enumerate}
        \item We present a fully controllable approach for generating photorealistic images of humans performing various activities.
        To the best of our knowledge, this is the first system capable of generating complex human interactions from appearance-based semantic segmentation label maps.
        Unlike other works in this domain~\cite{lassner2017generative,wang2018high}, our approach is fully automatic and does not rely on manually annotated data in the target domain in any step throughout the process.
        
        \item We demonstrate the effectiveness of our method by producing dense predictions, UV mapping and depth, on a complex multi-person scene.
        To the best of our knowledge, our work is the first to demonstrate accurate UV mapping of a model trained solely on synthetic data.
        
        \item We intend to release the dataset created using our pipeline, consisting of 100,000 fully-annotated images of multiple humans interacting in the CMU Panoptic environment -- the PanoSynth100K dataset.
        Furthermore, we intend to release the PSP model trained on this dataset that is able to auto-annotate real CMU images with appearance segmentation (human parsing) labels.
    \end{enumerate}

    \section{Related Work}
    The benefits of using synthetic data for training has encouraged the research community to explore various approaches for resolving the synthetic-to-real domain gap.
    One group of methods consists of enforcing invariance to the source domains at the feature level~\cite{tzeng2014deep,hoffman2016fcns,ganin2016domain,peng2018synthetic,yue2019domain}.
    Another leading group of methods, known as domain randomization~\cite{TobinFRSZA17,tremblay2018training,borrego2018applying,yue2019domain,khirodkar2019domain,prakash2019structured}, randomizes the appearance of objects of interest.
    Thus, the model is forced to develop invariance to their appearance (in particular to their appearance in the real domain) and to focus on their underlying geometric properties.
    Our work belongs to a third group of methods where the goal is to train a network capable of transforming the ``style'' of the synthetic image to that of a real one while retaining the general ``content'' of the scene.
    We, therefore, focus the remainder of this section on these methods.
    Domain adaptation at the pixel level is particularly appealing since it does not require any modifications to the existing training infrastructure.
    The adaptation result is intuitive to interpret and straightforward to assess.
    While not the only method for style transfer~\cite{gatys2016image,li2018closed}, GANs are at the heart of most popular approach for generating photorealistic images from synthetic ones~\cite{zhu2017unpaired,wang2018high,isola2017image,park2019semantic,bousmalis2017unsupervised,atapour2018real,chen2019learning}.
    The approaches vary in the ways the synthetic image is manipulated, the level of dependency on the target network and the level of control they provide to the user over the expected output.
    Style transformation is learned implicitly through adversarial training.
    To control the outcome, one needs to impose constraints on the process to steer it in the desired direction.
    Isola~\etal~\cite{isola2017image} constrained the process by matching pairs of images from the source and the target domains.
    This approach can lead to impressive results, though significantly limits the approach's applicability since obtaining such pairs is usually non-trivial.
    Zhu~\etal~\cite{zhu2017unpaired} replace the matching-pair assumption by cycle consistency constraints.
    This resulted in a wider scope of applicability and appealing results in several domains.
    However, the method does not provide mechanisms to control the output.
    Hoffmann~\etal~\cite{hoffmann2019learning} improve the above by injecting structural information to the process in a form of semantic segmentation maps, cyclically constrained by the output of a segmentation model applied to the generated image.
    This tightly couples the process with a specific task of semantic segmentation. 
    cGANs~\cite{cgan} are a powerful tool to incorporate semantic features into the domain adaptation process.
    Recent work has shown promising results in generating realistic images from semantic segmentation label maps~\cite{park2019semantic,wang2018high}.
    In particular, in pix2pixHD~\cite{wang2018high} the authors train a cGAN capable of both producing high-resolution images from semantic segmentation maps and allowing to control their content using instance segmentation labels and instance-level attributes.
    In a follow-up work~\cite{park2019semantic}, results are further improved by introducing the conditioning signal into intermediate layers rather than only at the input layer.
    More recent approaches~\cite{chen2019learning} utilize additional sources of the high-level semantic information freely available for the synthetic domain, \eg, depth, to improve the synthetic-to-real domain adaptation.

\section{Our Approach}
    In this section, we describe in detail the components of our data generation pipeline.
    In Section~\ref{sec:modeling_humans} we describe how we 
    render avatars in the virtual analog of the CMU Panoptic Dataset -- the PanoSynth dataset.
    In Section~\ref{sec:app_seg} we describe the training of a human parsing model on the virtual dataset and its application to auto-annotate frames from the panoptic dataset.
    In Section~\ref{sec:cgan} we use the auto-annotated CMU Panoptic dataset to train a cGAN model~\cite{park2019semantic} to generate photorealistic images.
    Finally, in Section~\ref{sec:dense_prediciton} we describe how we used the generate data of the cGAN to train a UV mapping and depth model.
    A description of the modeling of the CMU environment is provided in the supplementary material.

\subsection{Modeling Humans}
\label{sec:modeling_humans}
    \begin{figure}[tb]
    \begin{center}
    \begin{tabular}{cc}
    \includegraphics[width=5cm]{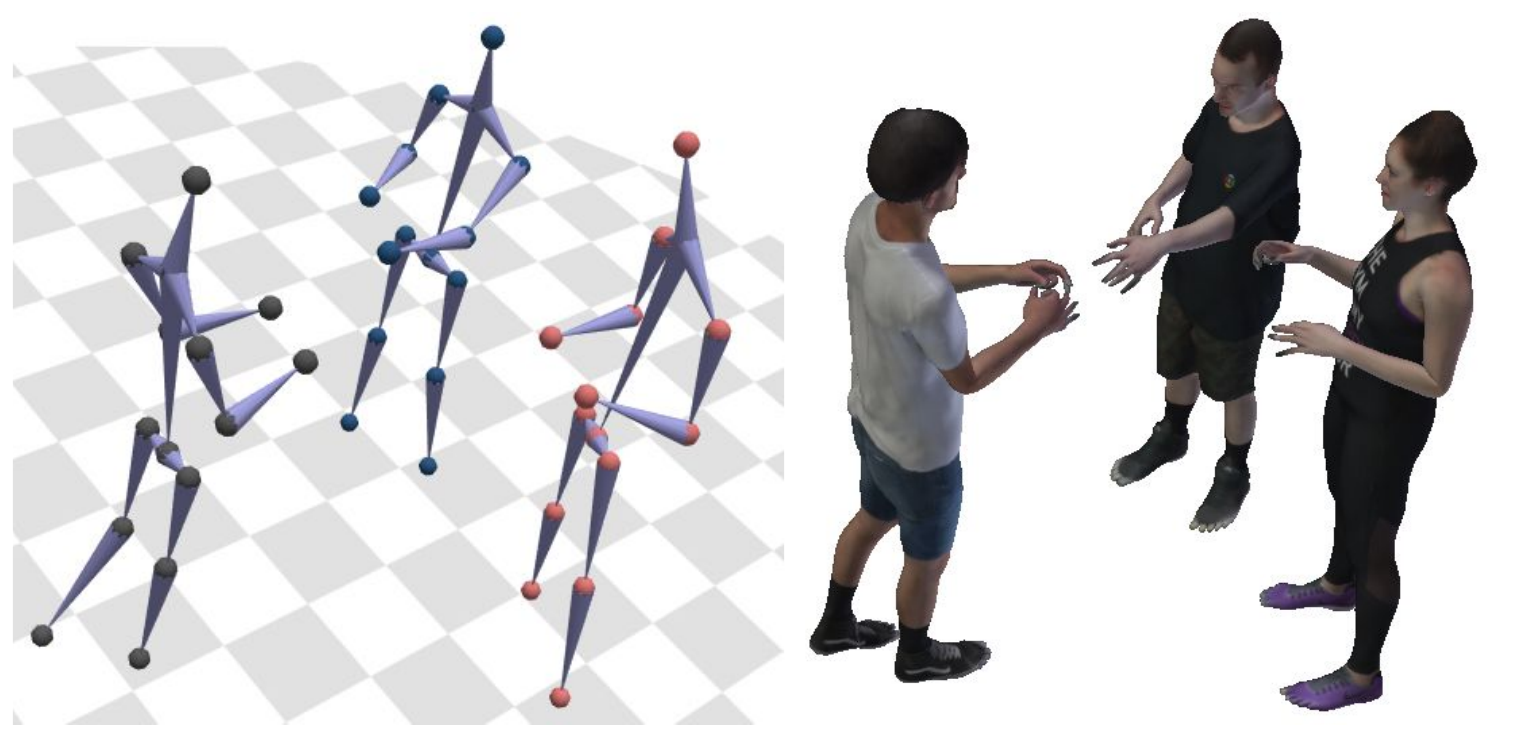} &
    \includegraphics[width=7cm]{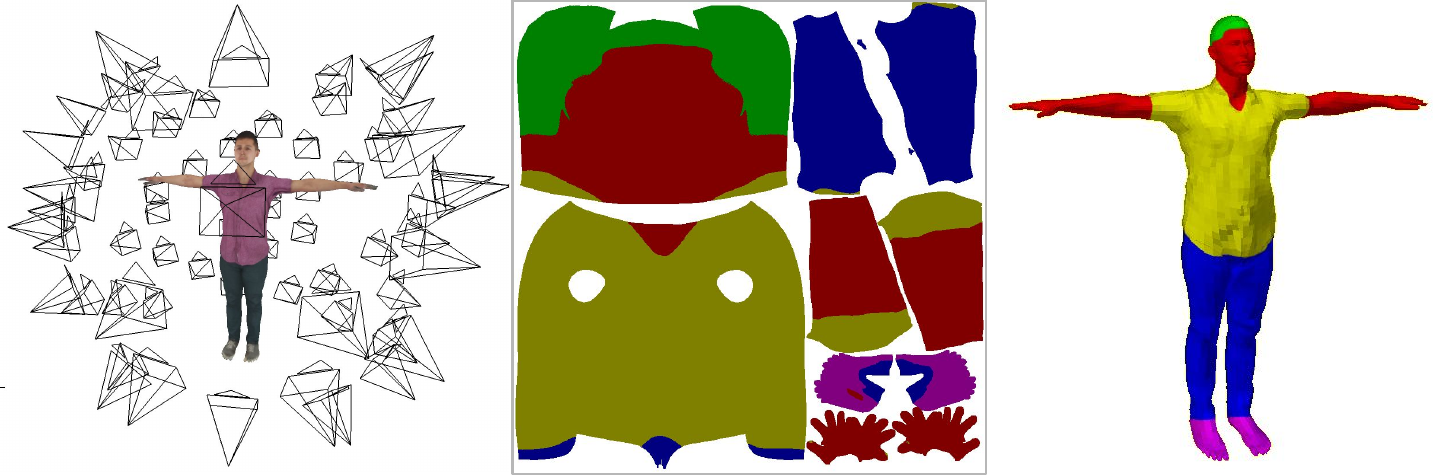}\\
    ~~~~~~~~~(a)~~~~~~~~~~~~~~~~~~~~~~~~~~~~~~~(b) & (c)~~~~~~~~~~~~~~~~~~~~~~~~~(d)~~~~~~~~~~~~~~~~~~~~~~~~~(e)
    \end{tabular}
    \caption{(a) Skeletons of humans provided by the CMU Panoptic Dataset (input to retargeting). (b) Synthetic avatars (retargeting output). (c) Different views used to scan the avatar in the ``virtual scanning'' procedure. (d) The full texture segmentation map after all of the views were aggregated. (e) The full labeling texture applied to the 3D avatar.}
    \label{fig:skeleton_to_avatar}
    \end{center}
    \end{figure}
    We represent human avatars in our framework using the SMPL model~\cite{loper2015smpl}.
    This model allows us to independently configure the avatar's pose and shape using two sets of parameters, $\bm{\beta} \in \mathbb{R}^{M}$ and $\mathbf{p}\in\mathbb{R}^{3\times 3\times N}$, for shape and pose, respectively.
    By using the SMPL canonical mesh topology, we can automatically label its texture maps using a method we label ``virtual scanning''.
    This provides us with automatically generated labels for the human parsing task (see Figures~\ref{fig:skeleton_to_avatar}(c-e)).
    A detailed description of this process is provided in the supplementary material.
    
    \subsubsection{From Skeleton to Avatar}
    Let $\left\{\mathbf{x}_t\right\}_{i=1}^T$ denote a sequence of $T$ poses comprising a motion pattern.
    Each pose is defined as a set of $N$ 3D joints' locations in space, \ie, $\mathbf{x}_t \in \mathbb R^{3\times N}$.
    The joints' locations may be recorded using motion capture (MOCAP) sensors or computed using triangulated keypoints detected in a set of calibrated cameras. 
    First, we recover the shape parameters of the SMPL avatar by fitting the same human shape to a set of $K$ randomly sampled poses,
    $\{\mathbf{x}_i\}_{i=1}^K$:
    \be
    \begin{split}
    \bm{\beta}^* = \argmin_{\bm{\beta}, \left\{\mathbf{p}_i\right\}_{i=1}^K} \sum_{i=1}^K{\Delta(\bm{\beta}, \mathbf{p}_i,\mathbf{x}_i)} + \alpha||\mathbf{\bm{\beta}}||_2^2
    &+ \gamma\sum_{i=1}^K {||\mathbf{p}_i||_2^2}, 
    \end{split}
    \ee
    where $\Delta(\bm{\beta}, \mathbf{p}_i,\mathbf{x}_i)$ measures the 3D location discrepancy of the joints' locations of the SMPL avatar parametrized by $\mathbf{\beta}$ and $\mathbf{p}$, and the corresponding observations.
    The role of the regularization parameters $\alpha$ and $\gamma$ is to enforce the shape and the pose priors, respectively.
    Then, we fix the shape parameters to $\bm{\beta}^*$ and recover a sequence of $T$ poses, $\left\{\mathbf{p}_t\right\}_{t=1}^T$, by consequently solving a set of $T$ minimization problems:
    \be
    \mathbf{p}_t = \argmin_{\mathbf{p}} \Delta(\bm{\beta}^*, \mathbf{p},\mathbf{x}_t) + \gamma||\mathbf{p}||_2^2.
    \ee
    Let $\mathbf{z} \in \mathbb{R}^{3\times N}$ denote the set of 3D joint locations induced by the SMPL model parametrized with 
    $\mathbf{\beta}$ and $\mathbf{p}$. 
    We define the discrepancy between the avatar's joint locations and real person's joint observations as:
    \be
    \Delta(\bm{\beta}, \mathbf{p},\mathbf{x}) = \sum_{i=1}^N||\mathbf{z}^i-\mathbf{x}^i||_2^2.
    \ee
    To speed-up computation and enforce temporal consistency, an optimization for pose $t$ is initialized with the solution obtained for pose $t-1$.
    The described process is similar to~\cite{joo2018total,loper2014mosh}, while the difference is that we optimize only for the location of the skeletal joints and not for avatar's mesh vertices.
    This results in lower accuracy in terms of mesh reconstruction, but with a benefit of significantly improved optimization speed.
    We note that if the recovery of the exact human shape is important, additional terms can be added to the optimization and the SMPL model could be replaced with a more detailed one~\cite{pavlakos2019expressive,romero2017embodied}.
    Figure~\ref{fig:skeleton_to_avatar} shows an example of the retargeting process applied to a specific scene from the CMU Panoptic Dataset.

    \begin{figure*}[tb]
    \begin{center}
    \includegraphics[trim={1.5cm 3cm 5cm 3.7cm},clip,width=10cm]{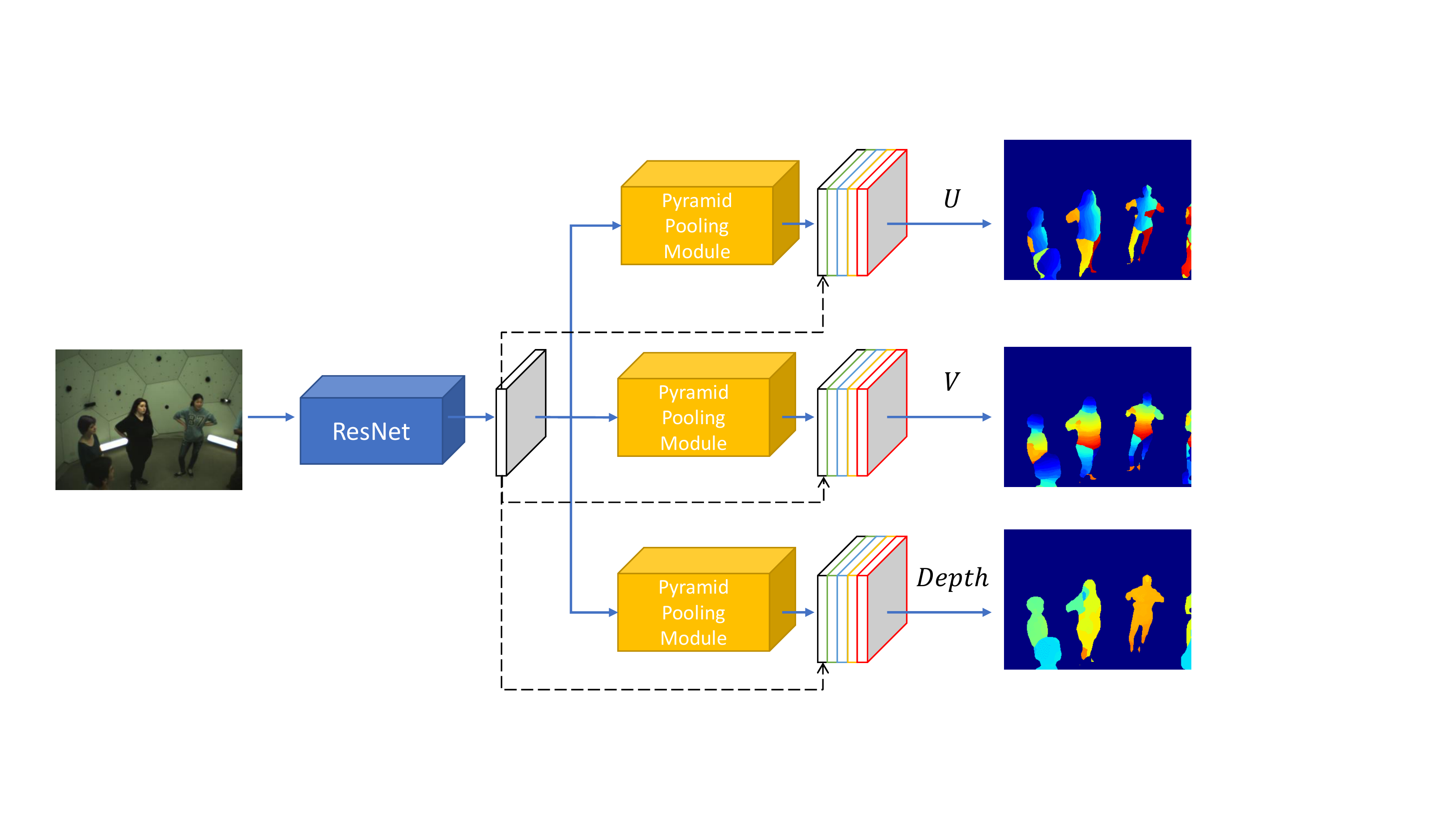} \\
    \caption{Illustration of our multi-head PSP network. The network is trained end-to-end as three segmentation tasks: U, V and depth.}
    \label{fig:multi_head_psp}
    \end{center}
    \end{figure*}

    \begin{figure*}[tb]
    \begin{center}
    \begin{tabular}{cc}
    \includegraphics[height=5.3cm]{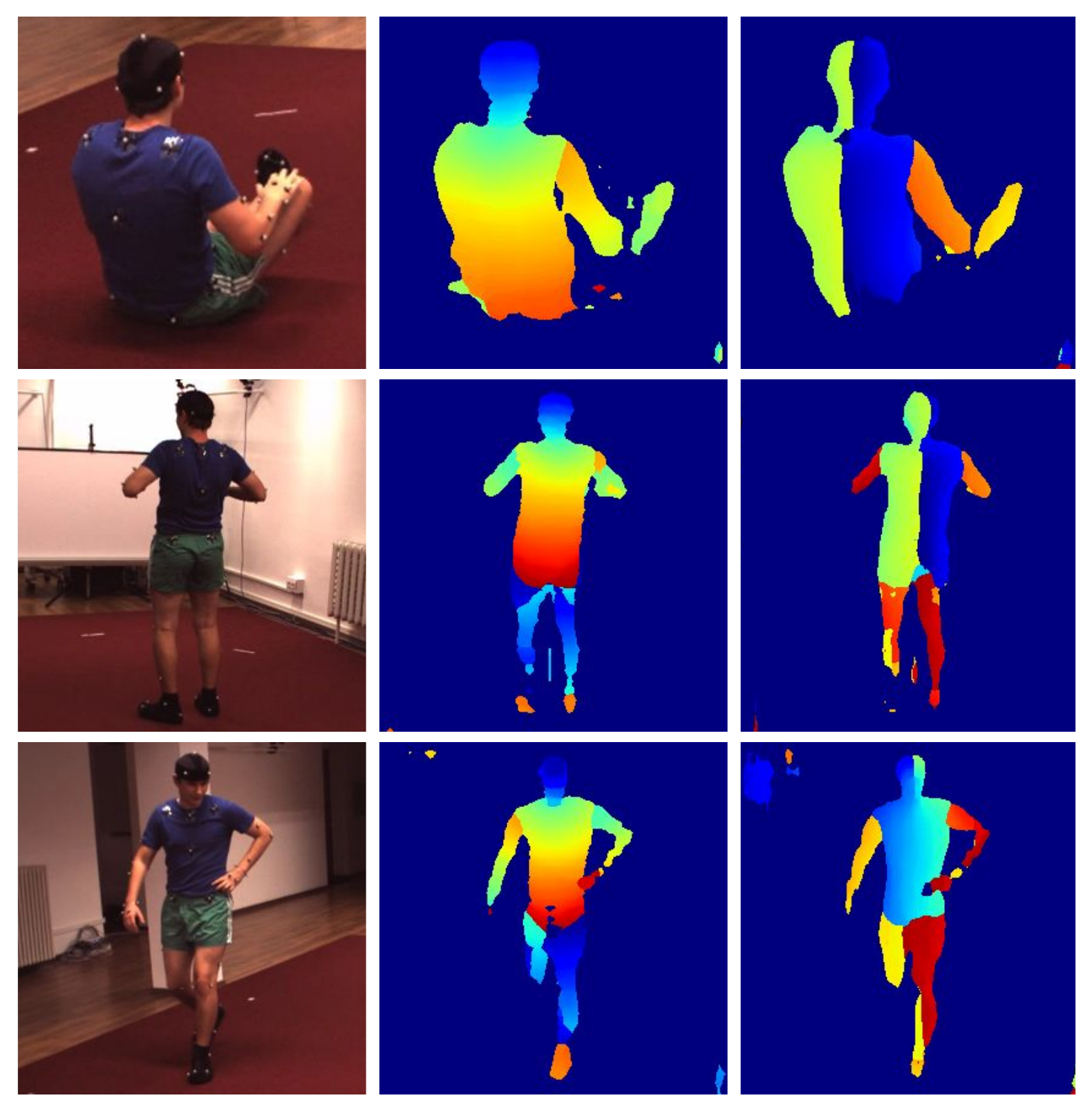} &
    \includegraphics[height=5.3cm]{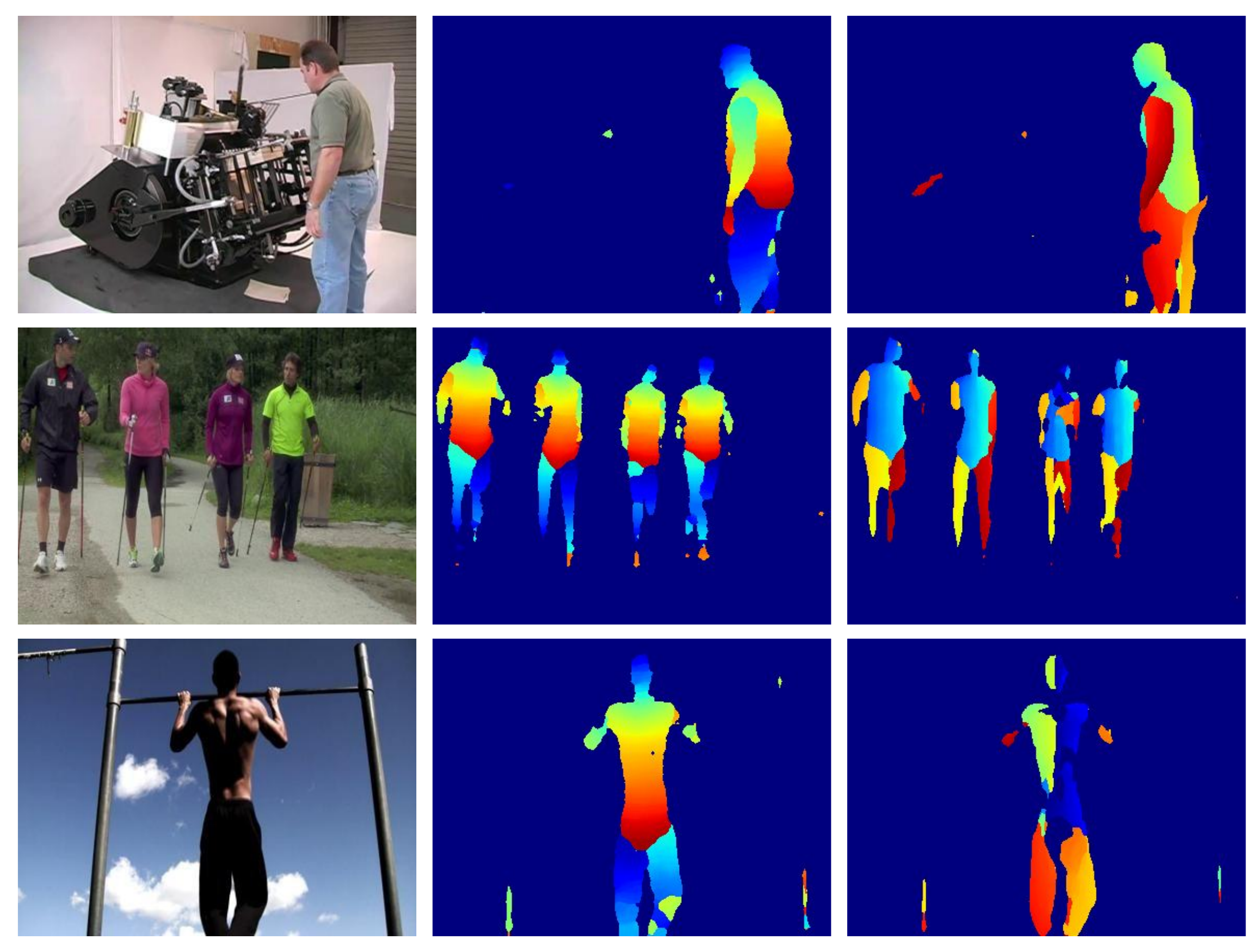}\\
    (a) & (b)
    \end{tabular}
    \caption{Selected images and UV mapping generated on the Human3.6 (a) and MPII Human pose (b) datasets. While our model was trained on images in a specific scenario, it was able to produce UV mappings on images from a different context.}
    \label{fig:mpi_and_human36}
    \end{center}
    \end{figure*}
    
\subsection{Human Parsing Model}
\label{sec:app_seg}
    In the domain of self-driving cars or scene parsing~\cite{cordts2016cityscapes,zhao2017pyramid} the object semantics is defined according to its type, \eg, car, pedestrian, river, sky, etc.
    In human parsing, different semantics are defined depending on the target application.
    For example, in applications involving pose estimation or body part detection, each pixel is associated with a body part label~\cite{varol17_surreal}.
    In the human parsing and fashion domain, additional labels, corresponding to different garment types are defined~\cite{ruan2019devil}.
    Such fine-grain semantic decomposition is good for training GANs on fashion datasets, as they contain a large variety of different garments.
    However, this label mapping is not suitable for training on the CMU Panoptic Dataset, where people usually wear casual clothing.
    Thus, we have redefined the semantics of appearance segmentation classes to better match the CMU Panoptic data and our intuition as to how we would like to generate human appearance.
    In particular, we define the set of labels to be generic enough allowing to describe the appearance of any individual, regardless of their dressing style.
    The classes are: skin, face, hair, top garment, bottom garment and shoes\footnote{Here we disregard the body parts' chirality.
    If required, any complementary information can be incorporated as additional segmentation labels.}.

    In the case of a single visible clothing article, \eg, dress, raincoat, it is labeled with the top garment label.
    In addition to the segmentation of humans, we add another annotation for the floor and the walls.
    We found that this was beneficial for training as it provided additional context.
    Our PanoSynth dataset contains 100,000 synthesized frames of different events containing multiple people posed as people from the original CMU Panoptic Dataset.
    We trained a Pyramid Scene Parsing (PSP) network~\cite{zhao2017pyramid} on the PanoSynth dataset and the Multi-Human Parsing (MHP) dataset~\cite{li2017towards}.
    We note that this dataset consists of 15K images, and is much smaller compared to our 100,000 samples synthetic dataset.
    Even though the MHP dataset is of another domain, we found that combining the two datasets allowed to overcome the domain gap applied to the original CMU Panoptic Dataset.
    The annotations predicted by the model described in this section were thereafter used to train the cGAN.
    
    \begin{figure*}[tb]
    \begin{center}
    \begin{tabular}{c}
    \includegraphics[width=\linewidth]{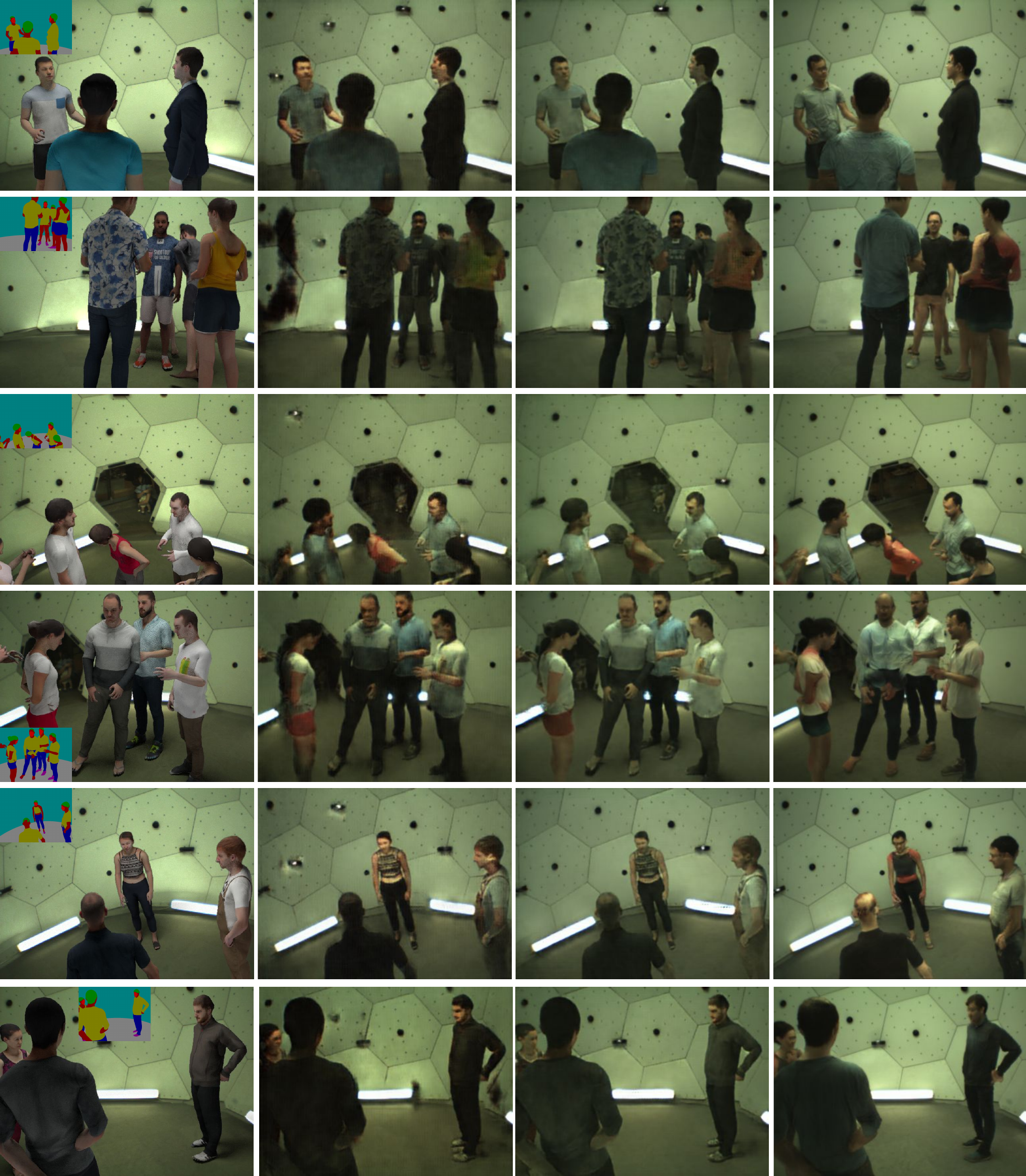} \\
   ~~~~(a)~~~~~~~~~~~~~~~~~~~~~~~~~~~~~~~~~(b)~~~~~~~~~~~~~~~~~~~~~~~~~~~~~~~~~(c)~~~~~~~~~~~~~~~~~~~~~~~~~~~~~~~~~(d)
    \end{tabular}
    \caption{Synthesized images transformed to be visually realistic.
    (a) Synthesized images. In the top-left corners of each image are the appearance segmentation maps (b) CycleGAN. (c) CyCADA. (d) Ours.}
    \label{fig:qualitative}
    \end{center}
    \end{figure*}    

\subsection{Conditional GAN}
\label{sec:cgan}
    We hypothesize that segmentation maps based on human parsing is more appropriate for our task than body part segmentation or instance segmentation.
    Instance segmentation maps are inadequate since human appearances are too varied for the GAN to be able to produce a coherent person from a person's silhouette.
    Body part segmentation may not be suitable since they do not necessarily correspond with the visual features of a person, \ie, a segmentation of a person's upper and lower arm may be misleading if they are wearing a shirt where the sleeve ends at a point other than the elbow.
    In initial experiments, the cGAN generated the foreground that were subsequently blended into the original background.
    This led to blending artifacts that impaired the images' authenticity, similarly reported in~\cite{dwibedi2017cut,alhaija2018geometric,tripathi2019learning}.
    Since we are interested in generating new images in only a limited set of  viewpoints, we resolve the blending issue by allowing the cGAN to re-create the background.
    To achieve this, we include a floor segmentation label.
    The floor's contour is a strong enough cue to allow the cGAN to ``memorize'' the background.
    Alternatively, we could have provided the camera ID explicitly or the background image itself.
    We use SPADE~\cite{park2019semantic} as the cGAN for this task with a minor modification.
    Park~\etal~\cite{park2019semantic} ``seed'' the generator by sampling from a Gaussian distribution.
    This method allows for the generation of diverse images given the same segmentation map.
    However, using this method there is no control over the generated person's appearance.
    To provide the cGAN with some control over the person’s appearance, we adopt the methodology proposed in~\cite{wang2018high} where we concatenate features produced by an autoencoder, averaged over instance segmentation maps.
    We note that the dimension of these features is deliberately very small (only three channels).
    This is to avoid giving the generator the entire information of the image.
    While the generator is still constrained to produce people that he saw during training, the inclusion of these maps provides guidance.
    We demonstrate the networks ability to produce diverse people in Figure~\ref{fig:same_frame}.
    These instance segmentation maps for the CMU dataset were produced by running an off-the-shelf Mask R-CNN model of the Detectron2~\cite{wu2019detectron2} framework.

\subsection{Dense Prediction}
\label{sec:dense_prediciton}
    To demonstrate the usefulness of our synthetic data we train a multi-task CNN to extract a UV mapping and a dense depth map for all the visible or partially visible people in the frame.
    We emphasize that acquiring ground-truth labels for these tasks is either very expensive, requiring dedicated hardware (depth cameras) or a rigorous annotation process~\cite{guler2018densepose}.
    For this task we implement a multi-head Pyramid Scene Parsing Network (MH-PSP).
    The network is constructed as three PSP networks with a shared ResNet152 backbone.
    See illustration in Figure~\ref{fig:multi_head_psp}.
    All three tasks were trained with cross entropy loss.
    The depth image dynamic range of 1 to 5.5 meters was uniformly quantized into 256 values.
    This leads to a nearly 2cm range per depth value.
    Similarly, the U and V maps were uniformly quantized to 255 values with an additional value reserved for background.
    We expect that including the residual regression unit of DenseReg~\cite{alp2017densereg} or combining the module with a Mask-RCNN~\cite{he2017mask,guler2018densepose} would improve the performance of the model.
    However, we found the simpler approach sufficient to demonstrate the synthetic data's usefulness.

\section{Experiments}
    In our experiments we build a dataset consisting of 16,176 synthesized images at 388x288 resolution.
    Rather than using the full 480 cameras available for the panoptic dataset, we constrain the images to 20 different camera views (one for each panel) that capture diverse views of the scene.
    We measure the quality of our pipeline by evaluating the generated images directly and by training a segmentation model on it and evaluating the model's performance on real data.
    In Section~\ref{sec:domain_gap} we measure the domain gap between the generated data and that of real data by two measures: the Fr\'echet Inception Distance (FID)~\cite{heusel2017gans} between them, and the impact on performance on a human parsing model trained on real data.
    In Section~\ref{sec:qualitative_analysis} we perform a qualitative analysis of the generated images.
    In Section~\ref{sec:multi_person} we train a new model on synthetic data to predict dense UV maps and depth.
    We evaluate the model's performance by measuring the accuracy of extracted 2D and 3D joints.
    In Section~\ref{sec:human_mpi} we evaluate the model qualitatively by demonstrating its performance on two external datasets: the MPII Human Pose Dataset~\cite{pishchulin16cvpr} and the Human3.6M Dataset~\cite{h36m_pami}.

\begin{table}    
    \centering
    \begin{tabular}{|l|c|c|c|}
    \hline
    Metric      & mIoU & mPixAcc & FID\\
    \hline\hline
    Synth+BG    & 0.60 & 0.83 & 123.4 \\
    \hline
    CycleGAN~\cite{zhu2017unpaired}    & 0.36 & 0.59 & 56.5 \\
    CyCADA~\cite{hoffman2017cycada}      & 0.44 & 0.65 & 36.8 \\
    Synth+BG+cGAN (Ours)        & \textbf{0.55} & \textbf{0.78} & \textbf{25.2} \\
    \hline
    \end{tabular}
    
    \caption{Comparison of our synthetic-to-real domain adaptation method to alternative approaches (best is highlighted). For FID, lower is better.
    }
    \label{table:baselines}
\end{table}    

\subsection{Measuring the Domain Gap}
\label{sec:domain_gap}
    We follow the quantitative evaluation methodology of~\cite{isola2017image,park2019semantic,wang2018high}, and measure mean intersection-over-union (mIoU) and mean pixel accuracy of the predictions of a model trained for semantic segmentation.
    Since the CMU Panoptic Dataset does not have segmentation maps available, we evaluate our performance using a semantic segmentation model (yet another PSP) trained on the MHP dataset and measure its performance on our synthesized images.
    The error of the model on the synthesized images can be decomposed into three factors: 1) model error -- the error the model achieves on its domain, 2) content gap between domains -- images of people in the wild vs. people in a specific context, and 3) appearance gap between domains -- images of real people vs. rendered images transformed by a GAN.
    The first two factors are roughly the same for all our baselines, as they are all applied to the same synthetic images.
    Therefore, this metric is an estimate of the third error -- the appearance gap.
    Additionally, we measure the FID between the transformed images and a subset of the real images from the CMU Panoptic Dataset.
    We applied our method to the rendered images, and compare them with the outputs of two well-known approaches for domain adaptation -- CycleGAN~\cite{zhu2017unpaired} and CyCADA~\cite{hoffman2017cycada}.
    Table~\ref{table:baselines} summarizes these results.
    
    A seemingly surprising result is the higher segmentation scores of the Synth+BG model.
    As apparent from Figure~\ref{fig:qualitative}, the human's textures in the purely synthetic images are very distinct, allowing for a trained model to easily segment out different appearance classes.
    Indeed, we take advantage of this as described in our ``virtual scanning'' procedure described in the supplementary material.
    As expected, when adding a semantic loss to the CycleGAN framework, \ie, CyCADA, the output images better adhere to these constraints, leading to better scores.
    The table suggests that replacing frameworks based on cycle-consistency with ours, based on dense correspondence of predicted label maps and images, preserves the most discriminability while also improving the most in realism, as measured by FID.
    The Synth+BG low FID score may seem counter-intuitive as the images are mostly comprised of pixels copied from real images (the background pixels), while other methods modify the background pixels, thus we would expect them to have a lower FID.
    We hypothesis that this did not happen due to two reasons: a) while transfer methods modify the backgrounds, they do so mildly.
    Furthermore, they may do so in a manner that does not affect the distribution of Inception activations. \eg, we do not expect a modest illumination difference to have such a large impact on the score.
    b) While the background pixels may cover roughly 60\% of the image, the activations of the inception network have a large receptive field that includes both foreground and background.
    Since the Inception model was pre-trained on ImageNet, we expect it to be more sensitive to people than to the simplistic background present in the Panoptic dataset.

\subsubsection{Qualitative analysis}
\label{sec:qualitative_analysis}
\begin{figure*}[tb]
    \begin{center}
    \begin{tabular}{cc}
    \includegraphics[width=6.4cm]{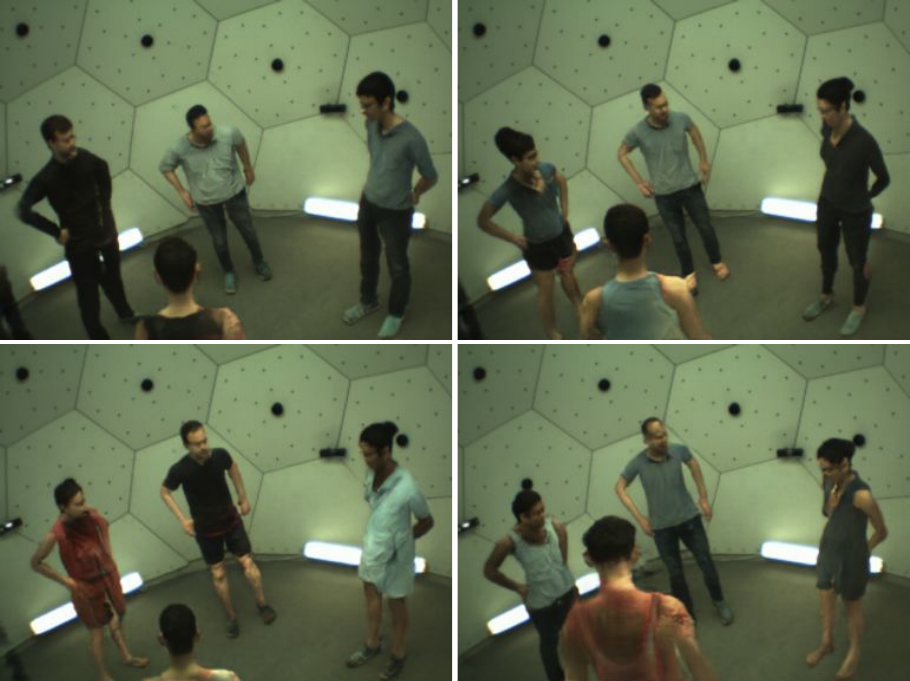} &
    \includegraphics[width=4.5cm]{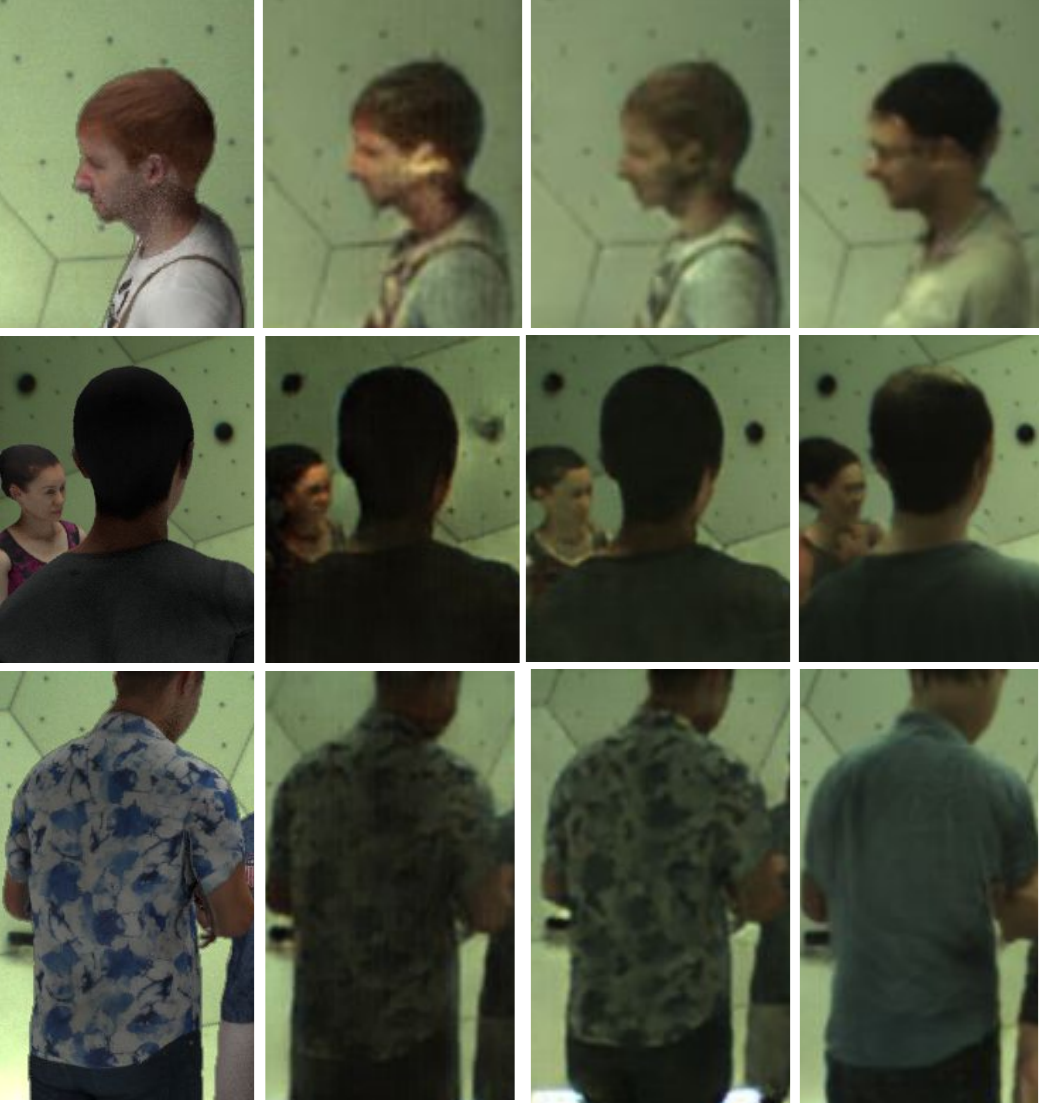}\\
    (a) & (b)
    \end{tabular}
    \caption{(a) Multiple generated images of the same scene with different avatars. (b) Close-ups on frames from Fig.~\ref{fig:qualitative} (from left to right: Rendered, CycleGAN, CyCADA, Ours).}
    \label{fig:same_frame}
    \end{center}
    \end{figure*}

    The main difference between CycleGAN and CyCADA and our method, is that the former two are trained to transform a synthetic image into a more realistic one, while the latter is trained to produce a realistic image from a segmentation map.
    This difference is well illustrated in Figure~\ref{fig:qualitative}.
    Note that columns (b) and (c), corresponding to the outputs of CycleGAN and CyCADA respectively, tightly follow the appearances of the input synthetic image.
    Column (d), corresponding to our method, preserves some features of the input image, such as the overall hue of the garments, but can vastly change other features such as skin tone and textures.
    Looking at the transformed backgrounds, note that both CycleGAN and CyCADA sometimes produce artifacts or hallucinate cameras, while these are not present in backgrounds generated by our model.
    Visually, we find the right-most column to be the most vivid.
    In Figure~\ref{fig:same_frame}(b) we include an enlarged portion images from Figure~\ref{fig:qualitative}.
    There are several imperfections in the rendering engine that are apparent in images that, in our subjective opinion, our method can overcome, while the baseline methods cannot.
    a) Realistic face: The synthesized images' faces are very distinct and unnaturally crisp.
    Our method produces plausible faces, correctly illuminated in accordance with the human's environment.
    b) The clothing in the synthetic images tends to be either smooth (as in rows 1 and 2) or overly detailed inappropriately to the context.
    Our method produces realistic folds and wrinkles in the images increasing their fidelity.
    In Figure~\ref{fig:same_frame} we demonstrate the flexibility of our model.
    Given a set of poses, we can produce photorealistic images of multiple humans with widely different appearances.
    This property may be useful as a form of data augmentation for scenarios posing a challenge in some human-related tasks.
    
\subsection{Multi-Person Scene Dense Prediction}
\label{sec:multi_person}
\begin{figure}[ht]
    \begin{center}
    \begin{tabular}{ccc}
    \includegraphics[width=4cm]{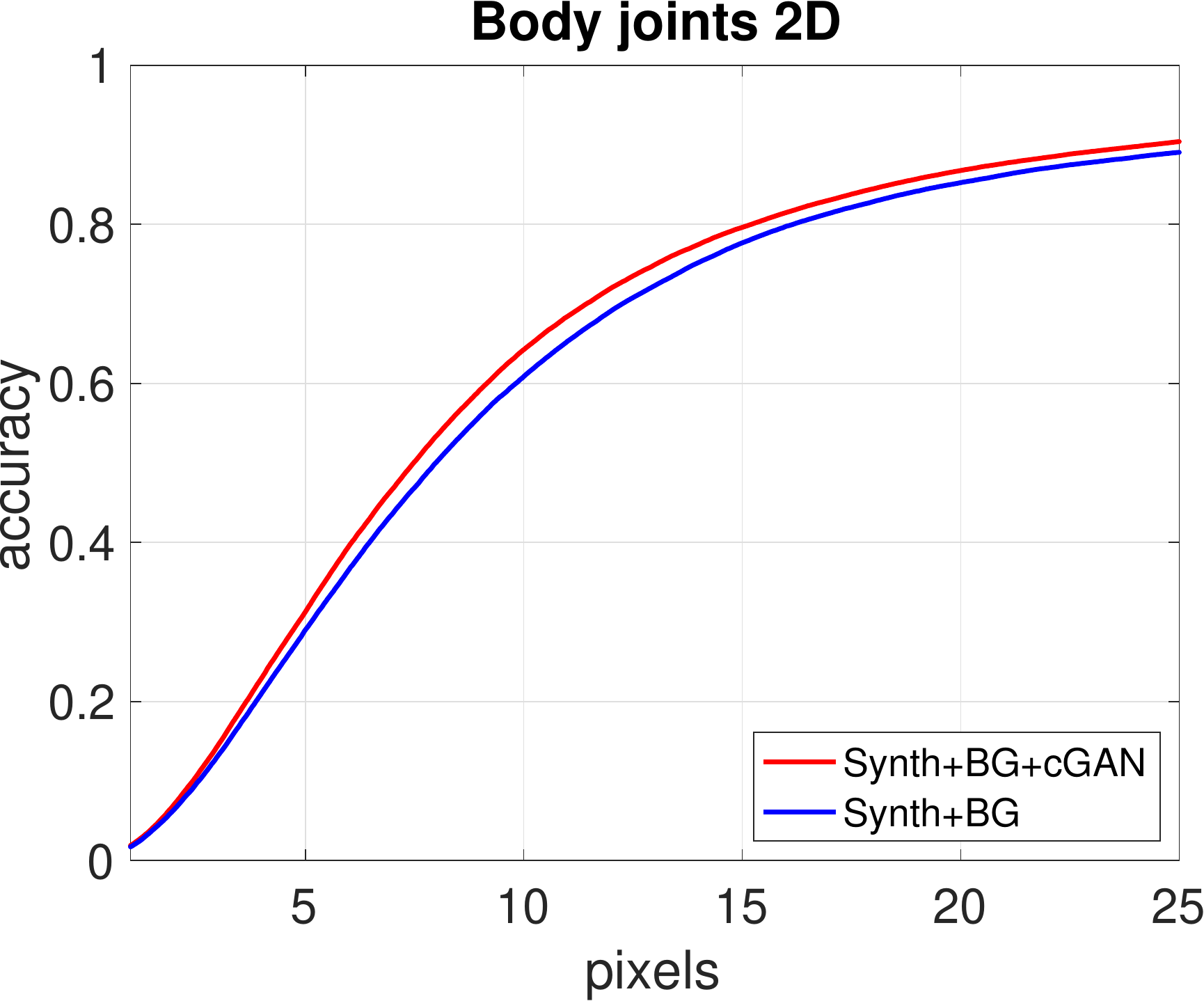} &
    \includegraphics[width=4cm]{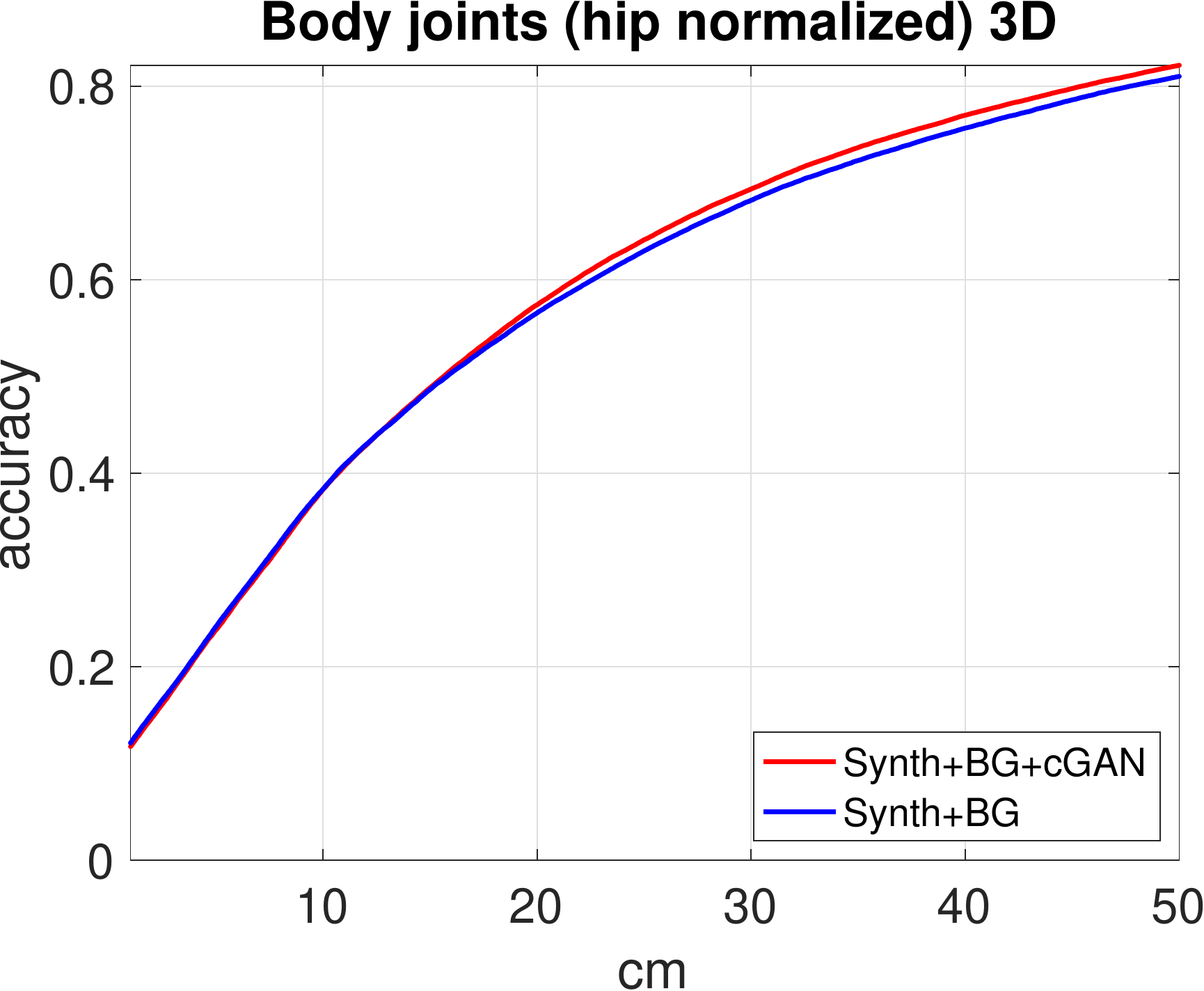} &
    \includegraphics[width=4cm]{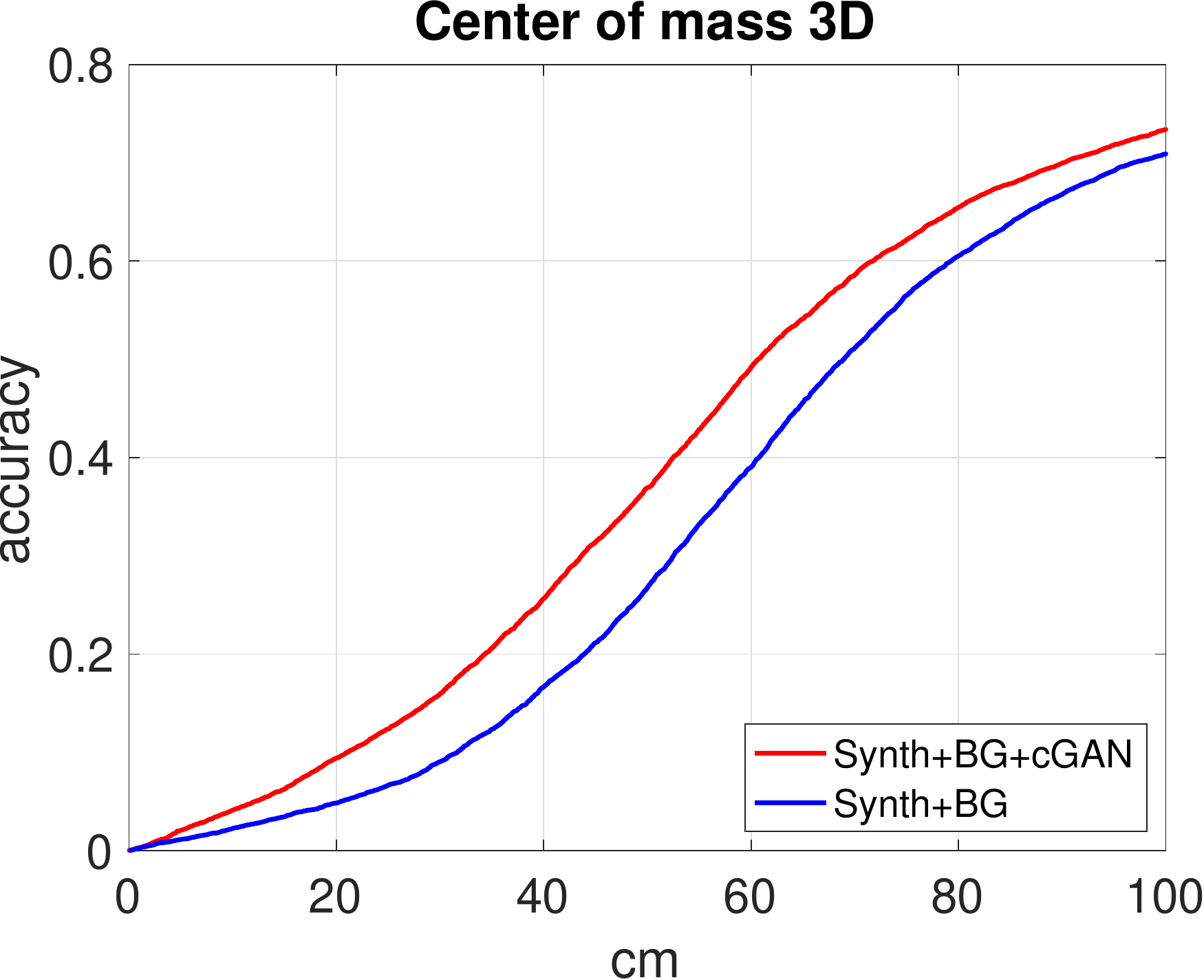} \\

    (a) & (b) & (c)
    \end{tabular}
    \caption{Accuracy of joint localization in 2D and 3D measured using the Probability of Correct Keypoint (PCK) metric. Overall, localization of  64,702 joints belonging to 13,285 skeletons were evaluated.
    The metrics are only calculated on correctly matched skeletons.
    The skeleton detection accuracies are 57\% and 62\% for Synth+BG and Synth+BG+cGAN respectively. Therefore, the results are skewed in favor of Synth+BG.
    (a) 2D joint detection (b) 3D joint depth accuracy relative to the center-of-mass (c) 3D skeleton center-of-mass absolute depth.}
    \label{fig:CMU_PCK_2D}
    \end{center}
    \end{figure}

    To validate the impact of the proposed domain adaptation technique, we train two models: before and after domain adaptation, previously denoted as Synth+BG and Synth+BG+cGAN.
    Since the CMU Panoptic Dataset does not have ground-truth UV mapping, we extract the 2D estimates for the body joint locations in the image plane from the inferred UV texture maps and use it as a metric to evaluate our model's performance.
    To extract 2D joints, we create a ``joint texture map'' where each joint is associated with a certain set of UV values.
    The texture is mapped to the UV prediction creating a image with joint predictions.
    To evaluate the quality of the depth maps, we project the joints' ground-truth locations onto the image plane and measure the distance between the estimated and ground-truth depth values.
    In other words, we measure the dense depth prediction using a sparse set of points for which the depth is given.
    We further divide the depth evaluation into skeleton center-of-mass absolute depth and joint depth relative to the center-of-mass.
    We note that the depth we measure is the distance to the body surface, while the ground truth is the depth to the actual skeletal joint, therefore this measure is intrinsically biased upward.
    This evaluation measures both the quality of the MH-PSP model and the quality of the data it was trained on.
    We emphasize that in this experiment we do not strive to reach state-of-the-art joint estimation accuracy, but to evaluate the quality of our UV mapping and depth maps.
    As ground-truth, we use the 3D skeletons, obtained using all 480 VGA cameras, projected to the image plane.
    As in~\cite{Joo_2017_TPAMI}, we use the Probability of Correct Keypoint (PCK) metric to evaluate the localization accuracy of body joints on the \emph{160422\_ultimatum1} sequence.
    Figure~\ref{fig:CMU_PCK_2D}(a) shows the performance as a function of the detection threshold (in pixels) for two models, trained on different types of synthetic data.
    Since our prediction is based on a monocular view and does not take into account a human body parametric model, it is not possible to predict occluded joints.
    Furthermore, in order to group joints and match them to a single person, we use instance segmentation maps produced by a Mask R-CNN network~\cite{he2017mask}.
    Thus, we report the localization accuracy only on joints which are visible in the frame.
    Therefore, we only measure the accuracy of detected joints.
    Figure~\ref{fig:CMU_PCK_2D} summarizes these results.
    The metrics are only calculated on correctly matched skeletons.
    The skeleton detection accuracies are 57\% and 62\% for Synth+BG and Synth+BG+cGAN respectively. This leads to a skew in the results in favor of Synth+BG.
    Thus, while the first two results demonstrate a mild improvement over the purely synthetic baseline, it is actually larger.

\subsubsection{Human3.6M and MPII Humans Pose Dataset}
\label{sec:human_mpi}
    Below we demonstrate the MH-PSP's performance on two datasets of different domains.
    The Human3.6M dataset focuses on pose estimation of a single person in a simple environment and the MPII dataset contains mostly single person images captured under challenging backgrounds.
    Although our approach focuses on a specific environment, the MH-PSP's predictions on these datasets demonstrate that the model indeed learned to capture features of a human rather than ``overfit'' to a specific scenario.
    Selected images are presented in Figure~\ref{fig:mpi_and_human36}.

\section{Discussion and Conclusions}
    In this work, we presented a method for generating photorealistic images of multiple people in a predefined context and demonstrated their usefulness for the tasks of UV mapping and depth prediction.
    By using human-centric synthetic data that was generated with a target context in mind, we were able to bridge the content gap to the extent that a small number of images from a different real domain were sufficient to overcome the synthetic-to-real appearance gap. 
    This work gives rise to potential future directions.
    While our generator was constrained only by the human parsing label map, it is possible to incorporate additional labels available with the synthetic data.
    Additionally, we may improve the degree of control of the cGAN by providing it with a richer and more descriptive set of features of the synthetic avatar.

\clearpage
%
%
\pagebreak
\appendix{\Large{\textbf{Supplementary Material}}}

\section{Virtual Scanning}
\begin{figure}[tb]
\begin{center}
\begin{tabular}{c}
\includegraphics[width=8cm]{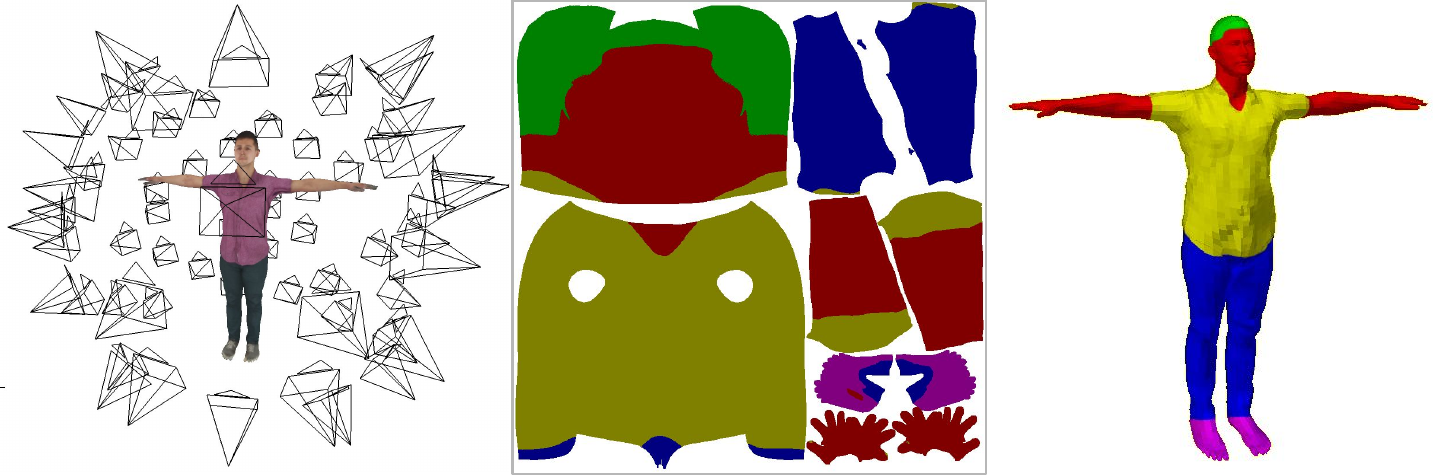} \\
(a)~~~~~~~~~~~~~~~~~~~~~~~~~(b)~~~~~~~~~~~~~~~~~~~~~~~~~(c)
\end{tabular}
\caption{(a) Different views used to scan the avatar. (b) The full texture segmentation map after all of the views were aggregated. (c) The full labeling texture applied to the 3D avatar.}
\label{fig:full_texture}
\end{center}
\end{figure}

To produce a large variety of appearances for synthetic humans we need a variety of 3D avatar textures labeled with appearance segmentation labels.
Here, again, manual annotation is not scalable and is likely to incur high costs.
We, therefore, automate this process by leveraging an off-the-shelf human parsing model~\cite{zhao2017pyramid} pre-trained on a dataset from the fashion domain~\cite{liang2015human}.
We render each one of the textured avatars in a T-pose from different views and apply our pre-trained human parsing model, resulting in a set of noisy label maps (see Figure~\ref{fig:full_texture}(a)).
The labeling imperfections are due to the two types of real-to-synthetic domain gaps: the appearance domain gap occurs since the human parsing model was trained on real fashion images while used on rendered avatars.
The content domain gap is due to the differences in scene geometry -- in the fashion domain, humans are typically captured from a frontal view, while to label the entire 3D body mesh we have to render from a variety of viewpoints.

\begin{figure}[tb]
\begin{center}
\begin{tabular}{c}
\includegraphics[width=8cm]{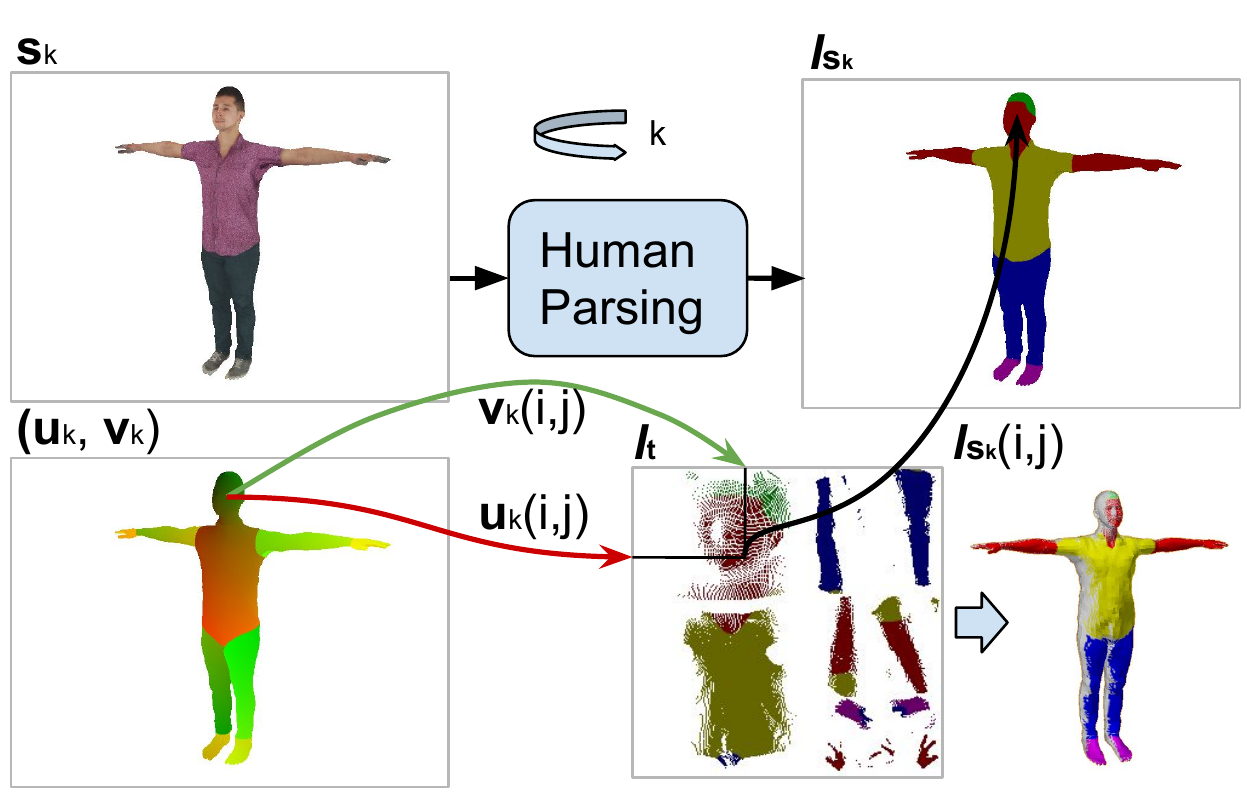}
\end{tabular}
\caption{Each pixel casts a label vote to the texture map. Note that the 3D avatar is only partially labeled from each single view. Fig~\ref{fig:full_texture} shows the final texture.}
\label{fig:virtual_scanner}
\end{center}
\end{figure}

The rendered avatar has a SMPL canonical mesh structure with a fixed UV mapping.
Therefore, each visible pixel in a rendered view casts a vote to a corresponding pixel in the texture map (see Figure~\ref{fig:virtual_scanner}).
The label of each pixel in the final texture is determined using a simple voting scheme. Figs.~\ref{fig:full_texture}(b) and (c) show an example of the final texture after aggregating all of the votes both as a texture map and as a rendered avatar, respectively.

More formally, let $\mathbf{s}$ denote a rendered RGB view of a SMPL avatar using texture map $\mathbf{t}$.
Let $(\mathbf{u}(i,j), \mathbf{v}(i,j))$ denote a UV texture mapping transforming a rendered pixel $(i, j)$ 
in $\mathbf{s}$ to its origin on the texture map $\mathbf{t}$.
Let $l_{\mathbf{s}}(i,j)$ denote an output of a given semantic segmentation model assigning label $l$, out of a set of possible labels $\mathcal{L}$, to pixel $(i,j)$ in $\mathbf{s}$. 
Thus, $l_{\mathbf{s}}$ provides a noisy labeling for $\mathbf{t}$: 
$l_{\mathbf{t}}(\mathbf{u}(i,j), \mathbf{v}(i,j))=l_{\mathbf{s}}(i,j)$. Given a set of $K$ rendered views with corresponding UV mappings, $\left\{\mathbf{s}_k, (\mathbf{u}_k, \mathbf{v}_k) \right\}_{k=1}^K$, we define the label of pixel $(u,v)$ in the texture map $\mathbf{t}$ as:
\be
l_{\mathbf{t}}(u, v) = \argmax_{l\in\mathcal{L}}  \sum_k \sum_{\overset{i,j:\mathbf{u}_k(i,j)=u\land}{~~\mathbf{v}_k(i,j)=v}} \mathbbm{1}\{l_{\mathbf{s}_k}(i,j)=l\}.
\ee
The attached video (\emph{VirtualScanning.mp4}) illustrates the virtual scanning process. Note, how the texture label map is being refined throughout the process. 

\section{Modeling the Environment}
\label{sec:modeling_env}
\begin{figure}[tb]
\begin{center}
\begin{tabular}{c}
\includegraphics[width=12cm]{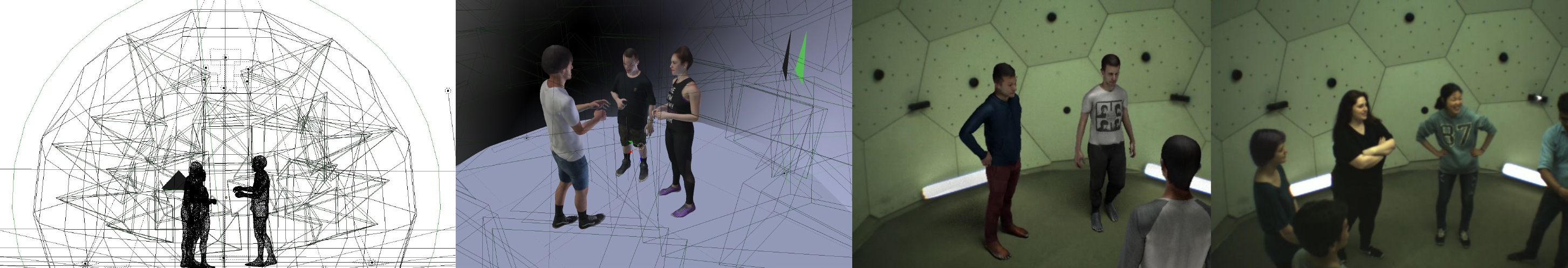} \\
(a)~~~~~~~~~~~~~~~~~~~~~~~~~(b)~~~~~~~~~~~~~~~~~~~~~~~~~(c)~~~~~~~~~~~~~~~~~~~~~~~~~(d)         
\end{tabular}
\caption{(a) Blender 3D model of the CMU Panoptic studio.  
         (b) Avatars inside the virtual dome. 
         (c) Example of a rendered frame.
         (d) Example of a real frame.}
\label{fig:VirtualDome}
\end{center}
\end{figure}

In order to reduce the content domain gap, it is important to incorporate as much of the prior knowledge on the target domain as possible. 
We use Blender\footnote{www.blender.org} to build the model of the environment and configure camera viewpoints corresponding to the real-world cameras.
This process can be automated given the architectural, illumination and camera layout plans of the target domain.
Figure~\ref{fig:VirtualDome} shows the VirtualDome - a modeled Panoptic Studio~\cite{Joo_2017_TPAMI} and examples of frames rendered from virtual viewpoints and real frames captured from corresponding cameras.

\subsubsection{Synthetic Human Interacting in Context}
\label{sec:human_in_context}

\begin{figure}[tb]
\begin{center}
\begin{tabular}{c}
\includegraphics[width=8cm]{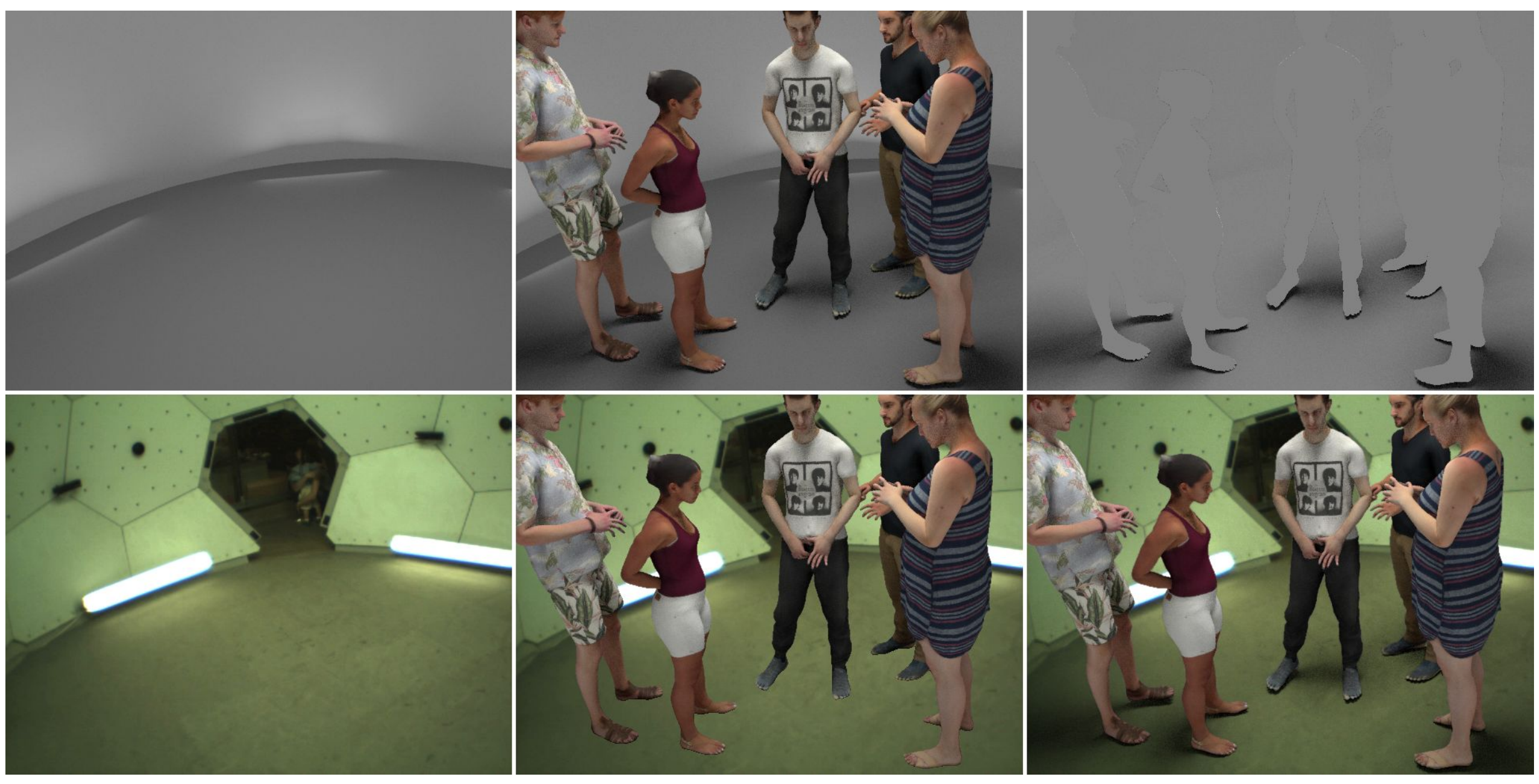} \\
(a)~~~~~~~~~~~~~~~~~~~~~~~~~(b)~~~~~~~~~~~~~~~~~~~~~~~~~(c)    
\end{tabular}
\caption{(a): Rendered (top) and real(bottom) background. 
         (b) Top - fully rendered scene. Bottom - real scene w/ alpha-blended rendered foreground. 
         (c) Top - ratio image (per pixel ratio between (b) and (a)) for background pixels.
         Bottom - real scene w/ alpha-blended rendered foreground and transferred shades.
         Note the improved photorealism w.r.t. (b).}
\label{fig:shading}
\end{center}
\end{figure}
Once the the environment and the humans participating in a certain event are modeled, we can place the virtual humans in the virtual environment.
Following the steps above, we can generate a large variety of fully annotated incarnations of an existing event or a novel event composed of different parts from existing ones (see Figure~\ref{fig:shading}(c) bottom).

To reduce the appearance domain gap for the background pixels, we exploit the tight correspondence between the modeled and the real environment as seen from camera views.
Given a synthetic image, we know the camera of which it was rendered and can associate it to the corresponding camera from the real world.
For this real camera, we can retrieve a typical background image.
Thereafter, we substitute the synthetic background pixels with the corresponding real image pixels. 
Figure~\ref{fig:shading}(b) top and bottom images show a rendered image with synthesized and real backgrounds, respectively.
Note that while the real background looks more natural, the lack of shadows increases the appearance domain gap of the entire frame.
To tackle this, we compute a ratio image~\cite{liu2001expressive} for the background pixels (top row of Figure~\ref{fig:shading}(c)), and apply it to the real background (bottom row of Figure~\ref{fig:shading}(c)).

Attached is a video (\emph{PanoSynth.mp4}) showing a sequence of generated frames together with different types of automatically generated annotations. Note that the final sequence (including the background and the shades) is perceived as more natural compared to the fully rendered one. You may also note that the fingers of the avatars are randomly articulated. The advantage of generating random articulation rather than a fixed one, \eg, an open palm, is that this reduces systematic biases in the data which may lead to overfitting of models trained on the data later on.

\bibliographystyle{splncs04}
\bibliography{PanoSynthArxiv}
\end{document}